\documentclass[letterpaper, 10 pt, conference]{ieeeconf}  

\IEEEoverridecommandlockouts                              

\usepackage{graphicx} 
\usepackage[utf8]{inputenc}
\usepackage{amsmath,amsfonts,booktabs,cite} 
\usepackage{gensymb} 
\usepackage{siunitx,textcomp} 
\usepackage[hidelinks]{hyperref}
\usepackage{cleveref} 
\usepackage{subcaption}
\usepackage{tabularx}
\usepackage{makecell}
\usepackage{pbox}
\let\labelindent\relax
\usepackage[inline]{enumitem} 
\usepackage[nolist]{acronym} 
\usepackage{multirow}
\usepackage[ruled,linesnumbered, vlined]{algorithm2e}
\usepackage{bm}
\usepackage{titlesec}
\usepackage{balance}
\usepackage[export]{adjustbox}
\usepackage{cellspace}

\usepackage[dvipsnames]{xcolor}

\usepackage[font=small, justification=justified]{caption}

\def\secref#1{Sec.~\ref{#1}}
\def\figref#1{{Fig.~\ref{#1}}}
\def\tabref#1{{Tab.~\ref{#1}}}
\def\eqref#1{Eq.~(\ref{#1})}

\DeclareMathOperator*{\argmin}{argmin}

\newcommand\normx[1]{\Vert#1\Vert}

\newcommand\etal{~\emph{et al. }}

\newsavebox{\twosubbox}

\crefname{algocf}{alg.}{algs.}
\Crefname{algocf}{Algorithm}{Algorithms}

\titleclass{\subsubsubsection}{straight}[\subsection]

\newcounter{subsubsubsection}[subsubsection]
\renewcommand\thesubsubsubsection{\thesubsubsection.\arabic{subsubsubsection}}

\titleformat{\subsubsubsection}
{\normalfont\normalsize\bfseries}{\thesubsubsubsection}{1em}{}
\titlespacing*{\subsubsubsection}
{0pt}{3.25ex plus 1ex minus .2ex}{1.5ex plus .2ex}

\makeatletter
\renewcommand\paragraph{\@startsection{paragraph}{5}{\z@}%
	{3.25ex \@plus1ex \@minus.2ex}%
	{-1em}%
	{\normalfont\normalsize\bfseries}}
\renewcommand\subparagraph{\@startsection{subparagraph}{6}{\parindent}%
	{3.25ex \@plus1ex \@minus .2ex}%
	{-1em}%
	{\normalfont\normalsize\bfseries}}
\def\toclevel@subsubsubsection{4}
\def\toclevel@paragraph{5}
\def\toclevel@paragraph{6}
\def\l@subsubsubsection{\@dottedtocline{4}{7em}{4em}}
\def\l@paragraph{\@dottedtocline{5}{10em}{5em}}
\def\l@subparagraph{\@dottedtocline{6}{14em}{6em}}
\makeatother

\title{\LARGE \bf
DawnIK: Decentralized Collision-Aware Inverse Kinematics Solver 

for Heterogeneous Multi-Arm Systems
}

\author{Salih Marangoz$^*$ \and Rohit Menon$^*$ \and Nils Dengler \and Maren Bennewitz
  \thanks{All authors are with the Humanoid Robots Lab, University of Bonn, Germany. Maren Bennewitz is additionally with the Lamarr Institute for Machine Learning and Artificial Intelligence, Germany. This work has partially been funded  by the Deutsche Forschungsgemeinschaft (DFG, German Research Foundation) under Germany’s Excellence Strategy, EXC-2070 -- 390732324 -- Phenorob and under grant 459376902 -- AID4Crops, and by the European Commission under grant agreement number 964854 -- RePAIR -- H2020-FETOPEN-2018-2020).}
  }

\begin{document}

\maketitle
\thispagestyle{empty} 
\pagestyle{empty}

\def\thefootnote{*}\footnotetext{These authors contributed equally to this work.} 

\begin{abstract} 
	Although inverse kinematics of serial manipulators is a well studied problem, challenges still exist in finding smooth feasible solutions that are also collision aware. 
	Furthermore, with collaborative service robots gaining traction, different robotic systems have to work in close proximity. 
	This means that the current inverse kinematics approaches do not have only to avoid collisions with themselves but also collisions with other robot arms. 
	Therefore, we present a novel approach to compute inverse kinematics for serial manipulators that take into account different constraints while trying to reach a desired end-effector pose that avoids collisions with themselves and other arms. 
	Unlike other constraint based approaches, we neither perform expensive inverse Jacobian computations nor do we require arms with redundant degrees of freedom. 
	Instead, we formulate different constraints as weighted cost functions to be optimized by a non-linear optimization solver. 
	Our approach is superior to the state-of-the-art CollisionIK in terms of collision avoidance in the presence of multiple arms in confined spaces with no  collisions occurring in all the experimental scenarios. 
	When the probability of collision is low, our approach shows better performance at trajectory tracking as well. 
	Additionally, our approach is capable of simultaneous yet decentralized control of multiple arms for trajectory tracking in intersecting workspace without any collisions.
	
\end{abstract} 

\section{Introduction}
\label{sec:intro}
\begin{figure}[t] 
	\centering
	\includegraphics[width=0.98\columnwidth]{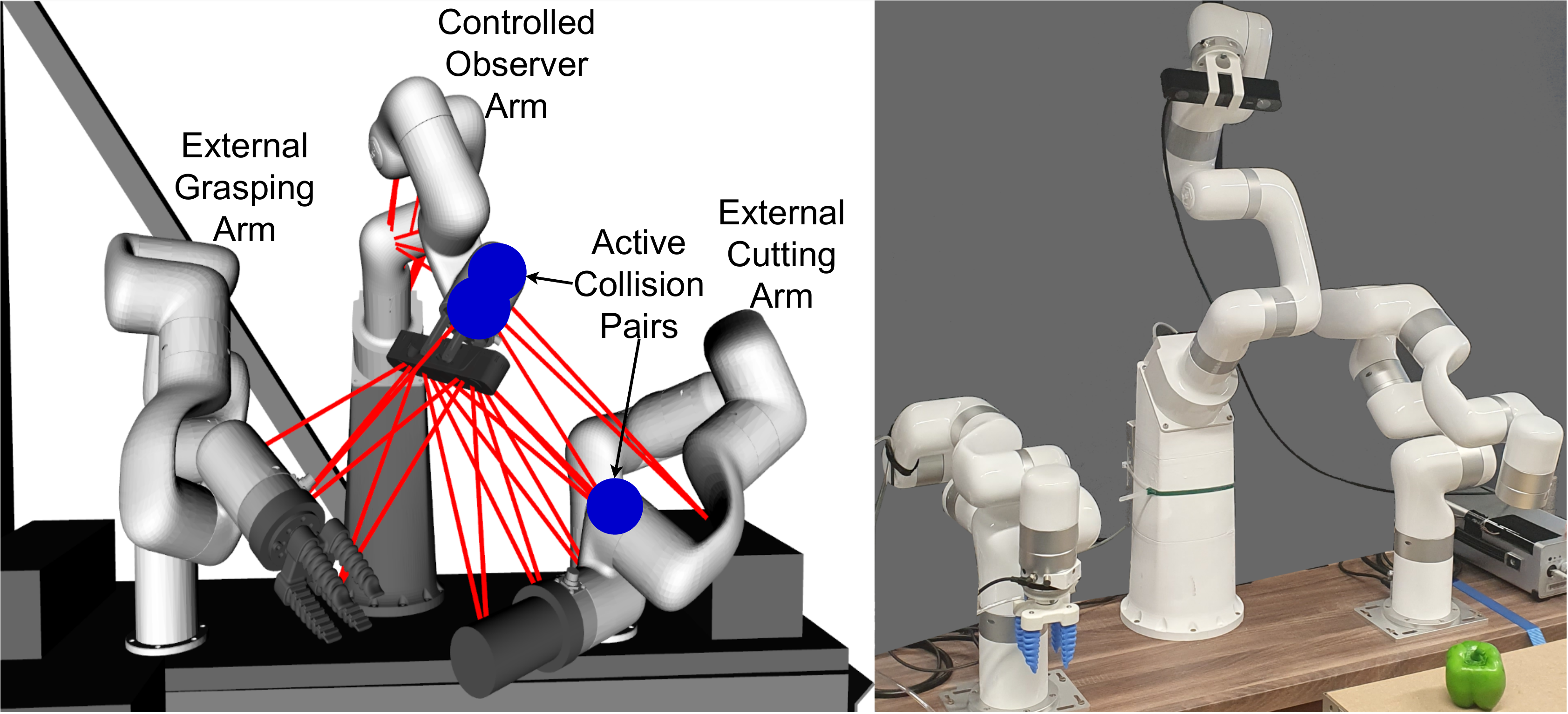} 	
	\captionsetup{width=0.99\columnwidth, justification=justified}
	\caption{Collision distance representation for HortiBot, a three arm system for active perception enabled robotic harvesting. 	
		Our proposed solver DawnIK enables control of the observer arm with collision-aware target following, while being aware of the current configuration of the externally controlled arms for grasping and cutting. 
		Red lines show the distance between the active collision pairs (exemplar shown in blue) taken into account for the inverse kinematics solution.} 
	\label{fig:cover_fig}
\end{figure}

With robots becoming ubiquitous, different arm based systems have to work in close proximity on the same task or related tasks concurrently. \textcolor{black}{These different systems may be controlled by software of different vendors and only their current joint positions and robot models are known, while no information regarding their task or their relative priorities is available.} 
Traditional approaches use workspace partitioning with virtual walls to avoid collisions between different arms or with humans \cite{takubo2002control}. 
However, this restricts the workspace and reduces the efficiency of the manipulation systems. 
At the other end of the spectrum, whole-body control approaches are used for humanoid manipulation planning~\cite{garcia2019integration, ferrari2017humanoid, sentis2006whole}.
These approaches are well suited for humanoids to maintain stability while simultaneously performing tasks
such as grasping, walking, and gaze stabilization
They are, however, sub-optimal for multi-arm systems that do not need to maintain coordination at all times. 
An example of this is the three-arm system HortiBot shown in \figref{fig:cover_fig}, which is added to the PATHoBot \cite{mccool21icra} for robotic harvesting of horticulture plants. 
The left and right arms are used for dual-arm manipulation for grasping and cutting respectively, whereas the observer or head arm, equipped with cameras is used for fruit mapping and tracking. 
Thus, the head arm needs to work independently during active fruit mapping, whereas during fruit harvesting it needs to provide the perception needed by the manipulation arms. 
In such scenarios, a collision-aware kinematics solver that can avoid collisions with the neighboring arms, static or moving, while following a specified end-effector path, is needed. 

In this paper, we present a novel non-linear optimization based approach called DawnIK to find an inverse kinematics solution for different Cartesian end-effector goals while avoiding self-collision and collision with other arms in the vicinity.
Our approach formulates different constraints including end-effector pose reaching as weighted cost functions with collision avoidance being the highest priority. 
The main contributions of our approach are as follows:
\begin{itemize}
	\item A decentralized approach to find collision-aware inverse kinematics where in addition to the controlled arm, externally controlled arms work in close proximity. 
	\item A method for fast and efficient collision distance calculation that scales well for multi-arm systems without using expensive perception algorithms.
	\item The capability to specify different end-effector constraints along with other constraints for singularity escape and kinodynamic limits.
\end{itemize}

Furthermore, we demonstrate through extensive experiments that our decentralized inverse kinematics solver can be applied to different multi-arm systems and outperforms an existing state-of-the-art inverse kinematic solver in terms of collision avoidance and trajectory tracking.

\section{Related Work}
\label{sec:related}
Robot arms are used to control the tool or sensor attached to the end-effector, i.e.,  the last link. 
As the end-effector needs six degrees of freedom to reach a desired pose in the SE(3) space,  most robot arms have six joints, with some having an extra seventh joint to provide kinematic redundancy. 

For effective use of robot arms, fast and reliable computation of inverse kinematics, i.e., calculating the set of joint values  that can place the end-effector at an arbitrary position and/or orientation in the arm's workspace, is critical.
However, this is still a challenging task and depends on the robot arm design\cite{paul1979kinematic, pieper1969kinematics}. 
Analytical, numerical, learning~\cite{malik2022deep, guo2019reinforcement}, and hybrid methods incorporating the former three, are the different categories of existing approaches\cite{aristidou2018inverse}. 
Although modern robot arms are designed such that closed form analytical solutions exist \cite{hawkins2013analytic} or can be constructed using code generation tools \cite{diankov2010automated}, they do not provide the flexibility for prioritization or relaxation of different constraints based on the task at hand. 

Typical numerical approaches use Jacobian methods \cite{buss2004introduction, buss2005selectively} to iteratively find the inverse kinematics solution for a desired pose around the current robot configuration. 
Inverse or pseudo-inverse Jacobian, or singular value decomposition (SVD) \cite{cao2023numerically} computations are expensive and especially prone to getting stuck in local minima. 
Recent approaches like TracIK\cite{beeson2015trac} and BioIK\cite{starke2020bio} have mitigated these problems to an extent using  random restarts with nonlinear optimization, and memetic algorithms that combine gradient-based optimization with genetic and particle swarm optimization, respectively. 
We tried to adapt BioIK\cite{starke2020bio} to include collision avoidance as a constraint.
However, with the constraint added, it suffered from a high failure rate in finding solutions.

As the above methods do not compute collision-aware trajectories, recent research has focused on collision-aware solvers.
Rakita\etal proposed RelaxedIK, a weighted-sum optimization method, that uses a neural network to approximate distances from collision states to find solutions that avoid self-collisions \cite{rakita2018relaxedik}. 
They also further extended it to CollisionIK to avoid collisions with environment obstacles using multi-objective constraint based optimization \cite{rakita2021collisionik}. 
Both RelaxedIK \cite{rakita2018relaxedik} and CollisionIK \cite{rakita2021collisionik} suffer from self-collisions due to the use of neural network approximation rather than actual distance computation. 
Also, in CollisionIK, only a maximum of three environment obstacles are considered and hence it is not suitable for avoiding another arm or arms moving in the vicinity. 

Approaches which exploit the null space of redundant manipulators for sequential solution of constraints have also been deployed in different systems. 
\textcolor{black}{Slotine \etal \cite{slotine1991general} presented a general framework for exploiting the kinematic redundancies of robot arms to fulfil functional and task specific constraints using recursive inverse Jacobian computations for task prioritization.}
Similarly, Zhao\etal\cite{zhao2021solving} proposed an approach for inverse kinematics of multiple redundant manipulators for collision avoidance using a combination of the inverse Jacobian and velocity obstacles. 
\textcolor{black}{However, these approaches are more suitable for highly redundant manipulators while at the same time being more computationally expensive due to expensive inverse Jacobian computations. 
We focus on finding inverse kinematics for manipulators in general which may not have redundant degrees of freedom.}

As our approach does not make any explicit use of redundancy, it can be applied in general to all manipulators and allows for the weighting of different goals to adapt to different tasks. 
More importantly, it provides a decentralized approach to collision-aware inverse kinematics computation for multiple arms for the first time.  
Additionally, our approach also does not depend on the expensive computation of inverse Jacobians for task priority resolution. 

\section{Problem Description}
\label{sec:problem_descr}
In this work, we consider the case where the motion of one or more arms are controlled in close proximity, but independent of, multiple other arms. 
The arm whose motion the inverse kinematics solver computes is referred to as the controlled arm. 
The arms that are controlled externally are active obstacles for the controlled arm, and are referred to as external arms.
In addition to the full access to the controlled arm's states, the solver has access to the robot model and current joint positions of the external arms.
Given a desired end-effector pose, we address the problem of computing the joint configuration of the controlled arm that can attain this end-effector pose while avoiding collisions with the external arms.

Formally put, given the controlled arm with with $n$ joints, whose current joint states are $ q^{\mathit{curr}}_i, i = 1...n$, current end-effector pose $ee^{\mathit{curr}}_{\mathit{pose}}$ and desired end-effector pose $ee^{\mathit{des}}_{\mathit{pose}}$, and neighboring robot arms with cumulative $m$ joints whose current joint states are $ q^{\mathit{curr}}_j, j = 1...m$,  the inverse kinematics solver finds a new set of joint commands $q^{\mathit{cmd}}_i$ for the controlled robot arm that satisfy the different constraints while being as close as possible to $ee^{\mathit{des}}_{\mathit{pose}}$.
We consider end-effector pose reaching as one of the constraints along with other constraints.

\section{Our Approach}
\label{sec:methods}
\begin{figure*}[t] \includegraphics[width=\linewidth]{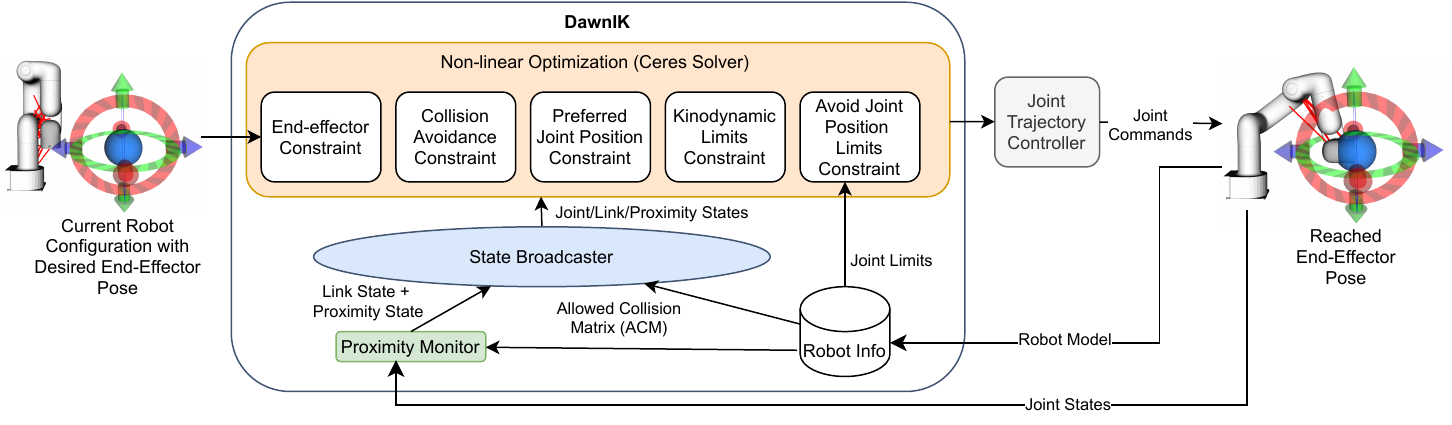} 	\captionsetup{justification=justified} 	\caption{Overview of the DawnIK solver.} 
	\label{fig:system_overview}
\end{figure*}
In this section, we present our approach for a decentralized collision-aware inverse kinematics solver based on Ceres\cite{agarwal2022ceres}, an open source C++ non-linear optimization solver. 
Our approach does not use perception to avoid collisions with other arms nor does it consider whole-body control approaches which view multi-arm systems as a single system for pose optimization. 
Instead, we address the problem of finding the inverse kinematics for the controlled arm, while being aware of the robot model and joint configurations of the external arms. 
Our approach consists of the robot parser module (\secref{subsec:robot_parser}) to compute offline static code from the robot description, the state broadcaster module (\secref{subsec:robot_state_br}) that computes the link states, collision distances, and command history, the proximity monitor (\secref{subsec:proximity}) that checks and computes the collisions distances, and the constraint solver module (\secref{subsec:optim}) that uses non-linear optimization to solve the objective function composed of the weighted cost functions formed from the different constraints.

\subsection{Robot Model Parser}
\label{subsec:robot_parser}
The robot parser module parses the robot description file and generates a header only file in C++ for the different robot variables: position, velocity and acceleration limits, link transformations, and allowed collision matrix. This allows for compile time optimization of the static transform calculation and other parameters. 
It also creates a simplified representation of the links using multiple spheres, as explained in \secref{subsec:proximity}.

\subsection{State Broadcaster}
\label{subsec:robot_state_br}
For the controlled arm as well as the obstacle arms, the states of all the links are calculated using the joint state information and the optimized robot model information generated by the parser module \secref{subsec:robot_parser}. 
Furthermore, the state of the collision objects for each link is computed.
For calculating the velocities, accelerations, and jerks, the command history for the last three iterations is stored.
The robot state, collision object information and command history are calculated and shared with all the objective functions in the solver module. 

\subsection{Proximity Monitor}
\label{subsec:proximity}
For collision-aware inverse kinematics, an efficient method for distance computation between different objects in the environment is essential. 
In general, there are three types of collision objects namely, the controlled arm's own links, the external arms' links, and other environment obstacles. 
In this work, although only the first two types are modeled, it is possible to add environment obstacles by fitting primitive shapes to the perception data.
The links of the controlled robot arm as well as the obstacle arms are modeled as a series of spheres. We use spheres as it is easy to calculate the collision distance. 
Furthermore, in addition to individual sphere surfaces being smooth, the spheres close to each other merge to provide a smooth differentiable surface, in comparison to boxes or capsules which have discontinuities at the boundaries.  

However, with an average of three spheres required to represent each link, the number of collision pairs can exponentially increase for multi-arm systems like the HortiBot \figref{fig:cover_fig}, leading to decrease in performance for the collision avoidance cost function. 
Hence, we first modify the allowed collision matrix (ACM), as for example provided by MoveIt\cite{coleman2014reducing}, that filters out collisions among external obstacle arms themselves in addition to adjacent or never colliding links of the controlled arm. 
Thus, collision checks are performed only for pairs that include the controlled arm's links, and any other collision is ignored.

The broad-phase collision manager provided by HPP-FCL\cite{montaut2022differentiable} is used for initial collision checking as it provides an efficient implementation of the GJK algorithm \cite{gilbert1988fast} for maintaining the bounding volume hierarchies. 
Furthermore, we use the axis-aligned bounding box (AABB) dynamic collision manager to filter out the collision objects that are not in close proximity to the controlled arm. 
The collision objects in proximity to the controlled arm are marked as active and the pairs are indicated using red lines in \figref{fig:cover_fig}.
For the active collision objects, the actual distance between the collision pairs using the link transformations derived from the joint states is computed and stored. 
The collision distances are then forwarded to the state broadcaster. 
\subsection{Constraint Solver}
\label{subsec:optim}
The inverse kinematics problem is formulated as a non-linear optimization problem with $p$ different constraints formulated as cost functions contributing to the overall objective function based on their respective weights. 
\begin{eqnarray}
	&\mathbf{q^{\mathit{cmd}}} = &\argmin_{q_i} \sum_{k=1}^{p}{ w_k \cdot f_k(q_i, \dot{q}_i, q_j, t)} \\
	&s.t. \hspace{0.1cm} &{q}^{l}_i < {q}_i < {q}^{u}_i
\end{eqnarray}
where $q_i$ and $\dot{q_i}$ represent the current joint position and velocities of the arm under consideration, $q_j$ represents the joint positions of the other arms, ${q}^{l}$ and ${q}^{u}$ represent lower and upper position limits, and $w_k$ represents the weight for the objective function $f_k$.
For each cost function $f_k$, the residuals $r_k$ are calculated first.
These residuals are then used to compute the loss function $\rho_{k}(s)$ where $s = \|r_k\|^{2}$.
For the individual cost functions, automatic differentiation (AutoDiff) is used to calculate the gradients. The non-linear optimization problem is solved using the trust region method\cite{conn2000trust} with the Levenberg-Marquardt algorithm\cite{levenberg1944method,marquardt1963algorithm}. 
The joint command variables are initialized using the current joint states $q^{\mathit{curr}}_i$. 
In addition, noise is added to the joint command variables to prevent the robotic arm from getting stuck in singularities like outstretched configurations.
The following different constraints are considered in our non-linear optimization framework.
\subsubsection{End-effector Constraint}
As stated earlier, the problem of reaching the desired end-effector pose is formulated as a cost function to the solver. The goal can be either the end-effector position, orientation or both. The capability to specify the type of goal is provided in the command being sent and the weights are adjusted accordingly. 
\begin{equation}
	\label{eq:ee_pose}
	\resizebox{.95\columnwidth}{!}
	{%
		$f_{\mathit{ee}} = w_{11}\cdot {f}^{\mathit{pos}} _{11}+  w_{12}\cdot {f}^{\mathit{orient}}_{12} \\
		\begin{cases}
		w_{12} = 0  & \text{position goal}  \\
		w_{11} = 0  & \text{orientation goal}   \\
		w_{11} \simeq  w_{12} & \text{pose goal}
		\end{cases}$
	}
\end{equation}
For example, if the task at hand needs only a specified position to be reached without orientation constraints, the current orientation is specified as the desired orientation and the weight $w_{11}$ associated with the position cost function is only specified whereas, $w_{12}$ associated with the orientation cost function is set to zero.

\subsubsection{Collision Avoidance Constraint} 
\begin{figure}[t] 
	\begin{subfigure}[t]{0.495\columnwidth} 		\centering
		\includegraphics[width=\columnwidth]{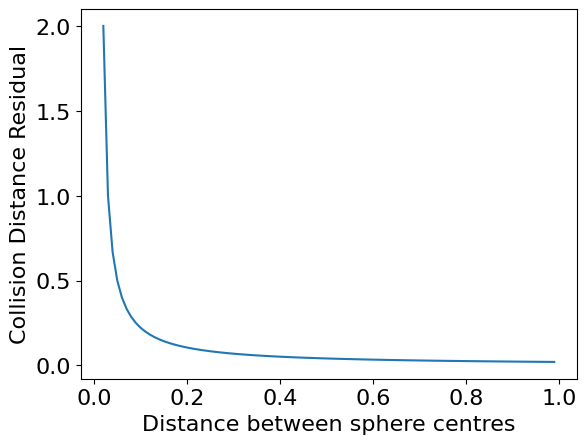} 	
		\captionsetup{width=0.9\columnwidth, justification=justified}
		\caption{Collision Distance Residuals for \eqref{eq:joint_limits} with $\epsilon_{\mathit{coll}} = 0.02$, $r_a = r_b = 0.05$.} 
		\label{fig:collision}
	\end{subfigure}
	\begin{subfigure}[t]{0.495\columnwidth} 		\centering
		\includegraphics[width=\columnwidth]{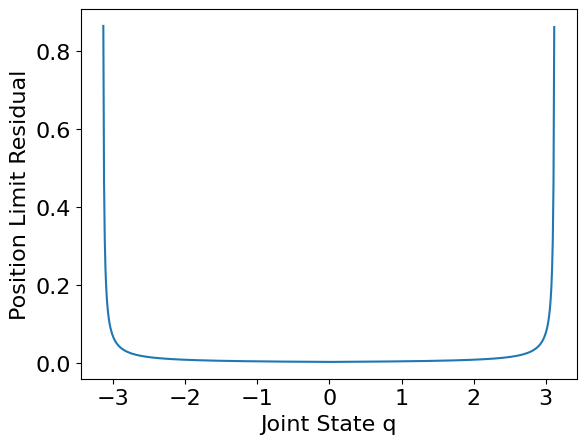} 	
		\captionsetup{width=0.9\columnwidth, justification=justified}
		\caption{Joint Position Limit Residuals for \eqref{eq:joint_limits} with $\epsilon = 0.01$, $q^{l}= -\pi$, $q^{u}=\pi$.} 
		\label{fig:joint_limits}
	\end{subfigure}
	\caption{Residuals for Avoid Joint Position Limits and Collision Constraints}
\end{figure}

The proximity monitor described in \secref{subsec:proximity} forwards all the active collision pairs to the solver via the state broadcaster. As the collision objects are modeled as spheres, the residuals for collision avoidance goal function is calculated as follows:
\begin{equation}
\label{eq:collision_residuals}
r^{\mathit{coll}}_{ab} = \frac{\epsilon_{\mathit{coll}}}{\normx{d_{ab} -r_a -r_b}} 
\end{equation} 
where $r_a$ and $r_b$ are the radii of the collision spheres and $d_{ab}$ is the distance between the centers. \figref{fig:collision} shows how the residuals increase inversely to the distance between spheres. 
\subsubsection{Preferred Joint Positions Constraint} 
Each joint $i$ has a preferred configuration $q^{\mathit{pref}}_i$ that can be updated during runtime.
Deviation of the joint states from the preferred joint positions is penalized using a Cauchy loss function.
In practice, for joints with position limits, we use the center joints, i.e., $\frac{q^{l}_i + q^{u}_i}{2}$ as the preferred joint positions.
\begin{equation}
f_{\mathit{pref}} = log(1+ \| r^{\mathit{pref}}_i \|)
\end{equation}
where $r^{\mathit{pref}}_i $ is the residual between the joint variables $q_i$ and the preferred joint positions $q^{\mathit{pref}}_i$.

\subsubsection{Avoid Joint Position Limits Constraint} 
A shortcoming of many inverse kinematics algorithms, e.g., Orocos KDL's methods \cite{smits_orocoskdl} is the inability to enforce joint position limits at each iteration.
In addition to setting the lower and upper parameter bounds, $q^{l}_i$ and $q^{u}_i$, we penalize the joint commands approaching close to the joint position limits as can been seen in \figref{fig:joint_limits}, by calculating the residuals as follows: 
\begin{eqnarray}
\label{eq:joint_limits}
r^{l}_i = \frac{\epsilon}{q_i - q^{l}_i - \epsilon}, \hspace{0.1cm}
r^{u}_i = \frac{\epsilon}{q_i - q^{u}_i - \epsilon} \\
f_{\mathit{poslim}} = \sum_{1}^{n}(\normx{r^{l}_i} + \normx{r^{u}_i})
\end{eqnarray}
where $\epsilon$ is a small non-zero number. 

\subsection{Kinodynamic Constraints}
The velocity, acceleration, and jerk that will be generated by the target joint position command $\mathbf{q^{\mathit{cmd}}}$  are calculated using backward differences  as shown in \eqref{eq:vel_acc_cmd}:
\begin{eqnarray}
\label{eq:vel_acc_cmd}
	\dot{q_t} = \frac{q_t - q_{\mathit{t-1}}}{\Delta t}, \hspace{0.2cm}
	\ddot{q_t} = \frac{\dot{q}_t - \dot{q}_{\mathit{t-1}}}{\Delta t} \hspace{0.2cm}
	\dddot{q_t} = \frac{\ddot{q}_t - \ddot{q}_{\mathit{t-1}}}{\Delta t}
\end{eqnarray}
The values  are filtered to minimize the noise caused by numeric differentiation. 
With a maximum jerk of 10 $rad/{seconds}^{3}$, we calculate the maximum permissible displacement in joint values and use that as the step size for the trust region.
Furthermore, the joint displacements are limited by calculating the residuals between the current joint states and new joint commands. 
Tukey loss, which aggressively attempts to suppress large errors, is used as the loss function.

\section{Experimental Evaluation}
\label{sec:exp}
The goal of our experimental setup is to provide a quantitative comparison of the performance of DawnIK with the state-of-the-art collision aware kinematics solver CollisionIK~\cite{rakita2021collisionik}. The evaluation is carried out with respect to trajectory tracking, collisions and singularities. 

\subsection{Experimental Setup}
\label{subsec:setup}
We carried out our experiments using the ROS-Noetic framework on a computer with core-i7 12700H processor, 32GB RAM and RTX3060 GPU. \textcolor{black}{The DawnIK solver runs at a frequency of 100 Hz for all the scenarios. }
\begin{figure*}[t]
	\begin{subfigure}[t]{0.33\linewidth} 		\centering
		\includegraphics[height=3.2cm]{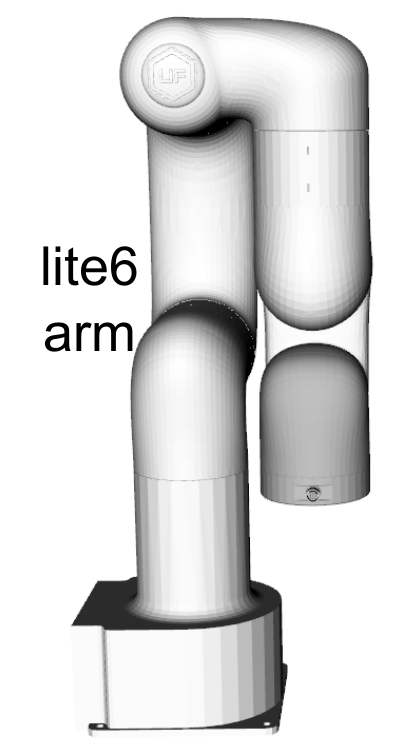} 	
		\captionsetup{width=0.9\columnwidth, justification=justified}
		\caption{Scenario 1} 
		\label{fig:scenario_1}
	\end{subfigure}
	\begin{subfigure}[t]{0.33\linewidth} 		\centering
		\includegraphics[height=3.2cm]{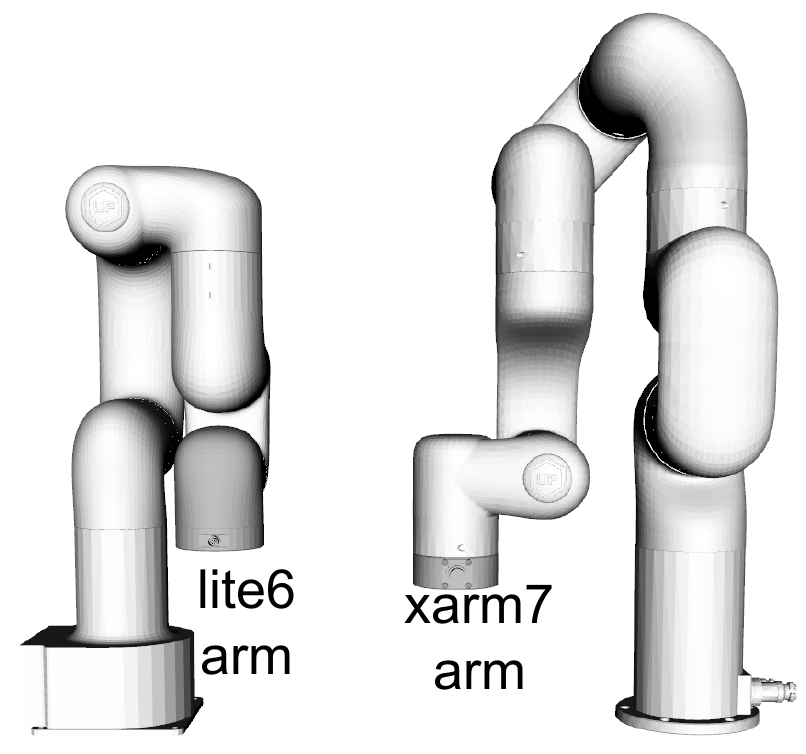} 	
		\captionsetup{width=0.9\columnwidth, justification=justified}
		\caption{Scenario 2 and Scenario 4} 
		\label{fig:scenario_2}
	\end{subfigure}
	\begin{subfigure}[t]{0.33\linewidth} 		\centering
		\includegraphics[height=3.2cm]{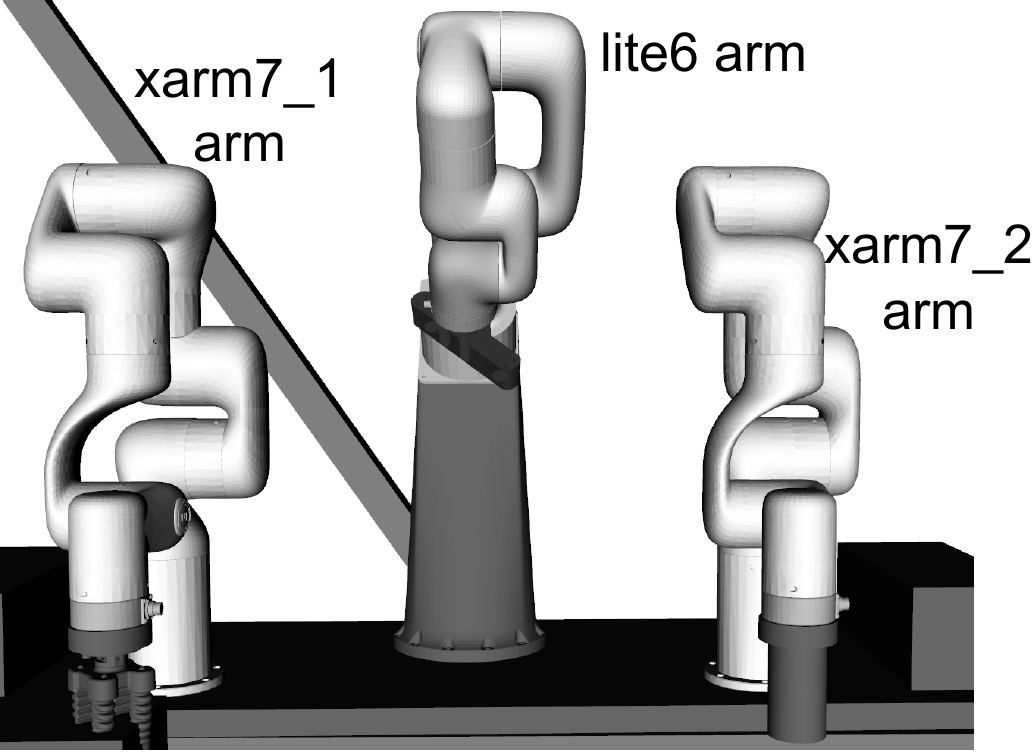} 	
		\captionsetup{width=0.9\columnwidth, justification=justified}
		\caption{Scenario 3} 
		\label{fig:scenario_3}
	\end{subfigure}
	\caption{Experimental Scenarios. 
	\textbf{Scenario 1}: Lite6 arm tested for self-collisions.
    \textbf{Scenario 2}: Lite6 controlled by DawnIK, Xarm7 controlled externally. Lite6 arm tested for self-collisions and collisions with Xarm7.
    \textbf{Scenario 3}: HortiBot Lite6 arm controlled by DawnIK, Xarm7s controlled externally. Lite6 tested for self-collisions ad collisions with both arms
    \textbf{Scenario 4}: Lite6 and Xarm7 controlled by DawnIK. Both tested for self-collisions and collisions with each other.}
	\captionsetup{justification=justified} 
	\label{fig:exp_scenarios}
\end{figure*}

We consider the following four scenarios for the quantitative analysis of our DawnIK solver.
\begin{itemize}
	\item \textbf{Scenario 1} \figref{fig:scenario_1}: Single 6 DoF Lite6 arm~\cite{ufactory2023Lite6Collaborative}, controlled by the IK solver and tested for self-collisions.
	\item \textbf{Scenario 2} \figref{fig:scenario_2}: Dual-arm system where the Lite6 is controlled by the IK solver and an Xarm7 arm ~\cite{ufactory2023xArmCollaborative} is controlled by MoveIt
	\item \textbf{Scenario 3} \figref{fig:scenario_3}: HortiBot system, where the Lite6 arm is controlled by the IK solver and the two Xarm7 arms are controlled by MoveIt
	\item \textbf{Scenario 4}: Same dual arm system as in Scenario 2, but both the Lite6 and Xarm7 arms are controlled by two independent DawnIK solvers. 
	We do not compare with CollisionIK for Scenario 4 as it was not possible to adapt it for controlling multiple arms.
\end{itemize}
For the experiments, the controlled arms have to follow different given paths, as defined below, in each scenario.
Furthermore, the external arms execute a given fixed trajectory repeatedly in Scenarios 2 and 3. 
Each experiment is repeated 5 times.
%

We use CollisionIK\cite{rakita2021collisionik} as our baseline, as it is the only other inverse kinematics solver capable of avoiding external collisions. 
We added the external arms as external collision objects and update their current poses to enable CollisionIK to consider them as dynamic external collision objects. 

The paths traced by the controlled arm are rigorously checked for collisions by using the fine mesh models of the arms. 
The trajectories traced out by the controlled arm and the external arms are recorded. 
Subsequently, each trajectory point is checked for collision using MoveIt's collision checking feature. 
Thus, we verify the effectiveness of the collision avoidance performed by both solvers.

\subsection{Results}
\begin{table*}[t] \centering
	\renewcommand{\arraystretch}{1.4}
	\begin{tabular}{|c|c|c|c|c|c|c|c|}\hline
		& x (mm)          & y (mm) & z (mm) & roll ($10^{-3}$rad) & pitch ($10^{-3}$rad) & yaw ($10^{-3}$rad)   & collisions\\ \hline
		S1-DawnIK      &\textbf{2.06 $\pm$ 5.60}               &5.93 $\pm$ 20.44  &\textbf{2.95 $\pm$ 9.59}  &0.27 $\pm$ 0.71  &0.11 $\pm$ 0.14  &\textbf{0.36 $\pm$ 1.03}   &0\\ \hline
		S1-CollisionIK &6.26 $\pm$ 9.37   &11.61 $\pm$ 22.88  &5.66 $\pm$ 12.07  &0.64 $\pm$ 1.15  &0.26 $\pm$ 0.46  & 0.92 $\pm$ 1.48  &0\\ \hline \hline
		S2-DawnIK      &20.06 $\pm$ 20.45                &18.01 $\pm$ 19.18  &12.43 $\pm$ 24.42  &0.92 $\pm$ 2.92  &0.58 $\pm$ 2.35  &\textbf{1.32 $\pm$ 1.78}   &0\\ \hline
		S2-CollisionIK &32.84 $\pm$ 39.50                &37.08 $\pm$ 48.40  &34.81 $\pm$ 55.50  &1.20 $\pm$ 1.73  &0.36 $\pm$ 0.68  &1.46 $\pm$ 1.78   &0\\ \hline \hline
		S3-DawnIK      &99.4 $\pm$ 134.65                &45.59 $\pm$ 57.34  &68.43 $\pm$ 79.22  &30.52 $\pm$ 55.43  &8.28 $\pm$ 17.76  &29.2 $\pm$ 58.01   &\textbf{0}\\ \hline
		S3-CollisionIK &27.48 $\pm$ 26.45                &39.46 $\pm$ 40.91  &31.10 $\pm$ 47.94  &6.208 $\pm$ 7.53  &0.981 $\pm$ 1.51  &4.98 $\pm$ 5.4   &38.13 \\ \hline
	\end{tabular}
	\captionsetup{width=0.97\linewidth, justification=justified}
	\caption{DawnIK versus CollisionIK mean end-effector errors with standard error, and mean number of collisions for the three different scenarios. S1, S2, and S3 indicate Scenarios 1, 2 and 3 respectively. The results demonstrate the ability of DawnIK to avoid collisions in all the three scenarios while being superior at trajectory tracking in Scenarios 1 and 2. Figures shown in bold indicate statistically significantly lower errors for p $<$ 0.05 using the Mann-Whitney U test. Note that in S3, no collisions occur using DawnIK solver whereas CollisionIK solutions lead to multiple collisions.}
	\label{tab:results}
\end{table*}
\subsubsection{Scenario 1}
\begin{figure*}[t]
	\begin{subfigure}[t]{0.33\linewidth} 		\centering
		\includegraphics[width=\columnwidth]{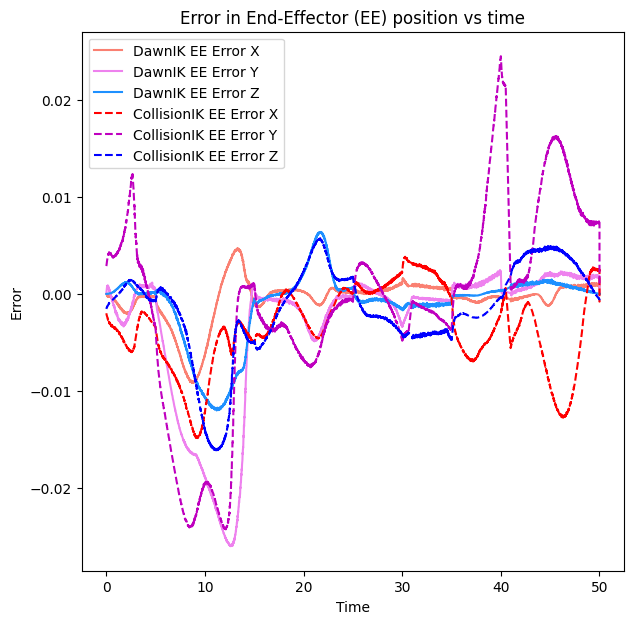} 	
		\captionsetup{width=0.97\columnwidth, justification=justified}
		\caption{Mean error in end-effector positions for DawnIK (dashed lines) and CollisionIK (solid lines) for all the trials in Scenario 1.} 
		\label{fig:ee_error}
	\end{subfigure}
	\begin{subfigure}[t]{0.33\linewidth} 		\centering
		\includegraphics[width=\columnwidth]{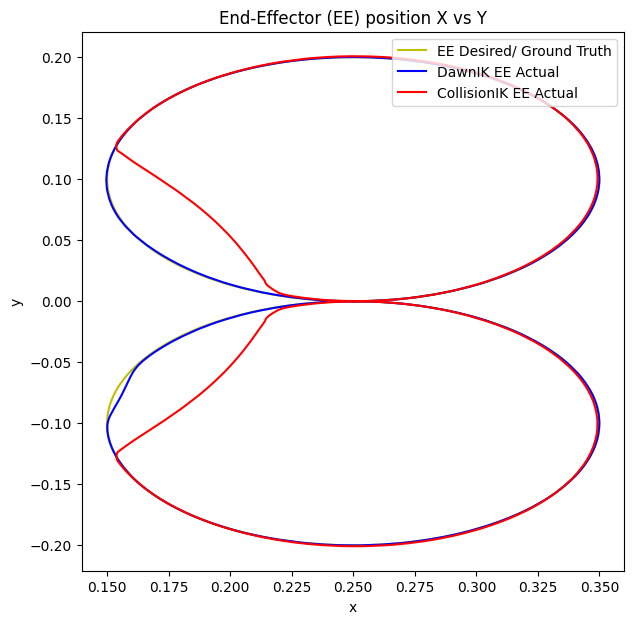} 	
		\captionsetup{width=0.97\columnwidth, justification=justified}
		\caption{Mean path traced by the end-effector using DawnIK (blue) and CollisionIK (red) while drawing an eight in the X-Y plane for Scenario 1.} 
		\label{fig:eight_xy}
	\end{subfigure}
	\begin{subfigure}[t]{0.33\linewidth} 		\centering
		\includegraphics[width=\columnwidth]{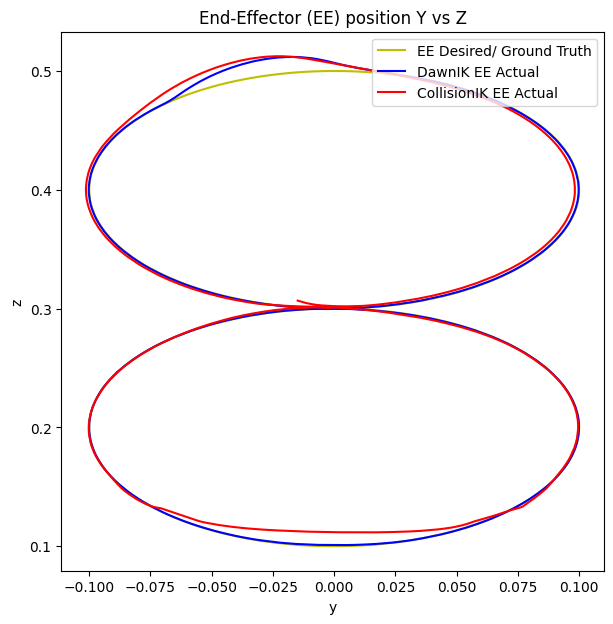} 	
		\captionsetup{width=0.97\columnwidth, justification=justified}
		\caption{Mean path traced by the end-effector using DawnIK (blue) and CollisionIK (red) while drawing an eight in the Y-Z plane for Scenario 1.} 
		\label{fig:eight_yz}
	\end{subfigure}
	\caption{Comparison of performance of DawnIK and CollisionIK in Scenario 1. \figref{fig:ee_error} show that DawnIK leads to lower end-effector position errors over time. \figref{fig:eight_xy} and \figref{fig:eight_yz} plots provide a visual demonstration of DawnIK's superior trajectory tracking.}
	\captionsetup{justification=justified} 
	\label{fig:scenario_1_plots}
\end{figure*}
In the first scenario, we compare the performance of DawnIK with CollisionIK for path tracing, where self-collisions can occur, using the Lite6 arm. 
For this and the other experimental scenarios, we generated three different path shapes (squares, circles, and the figure eight) in the X-Y
and Y-Z planes to trace them with the controlled arm.
The squares are of width 40 cm and the circles are of radius 18cm. The loops for the eight are circles of radius 10cm. 
\textcolor{black}{As both DawnIK and CollisionIK are inverse kinematic solvers and not trajectory planners, the waypoints for these trajectories are precomputed at a sampling rate of 30 ms and fed to both the solvers.}

As can be seen from \tabref{tab:results}, both DawnIK and CollisionIK avoid collisions for all the experiments in Scenario 1.
However, as can be seen visually from \figref{fig:ee_error} and the values for the end-effector position and orientation errors in \tabref{tab:results}, DawnIK outperforms CollisionIK in terms of trajectory tracking. 
We calculated the mean trajectories over 5 iterations for the task of drawing the figure in X-Y and Y-Z plane and plot the curves in \figref{fig:eight_xy} and \figref{fig:eight_yz} respectively. 
When there is a probability for self-collision, CollisionIK deviates from the path noticeably whereas DawnIK deviates only to a minimum extent. 
This is due to the fact that DawnIK uses actual distances to compute the collision avoidance constraint, and CollisionIK approximates it using a neural network.

\subsubsection{Scenario 2}
\begin{figure*}[t]
	\begin{subfigure}[t]{0.33\linewidth} 		\centering
		\includegraphics[width=\columnwidth]{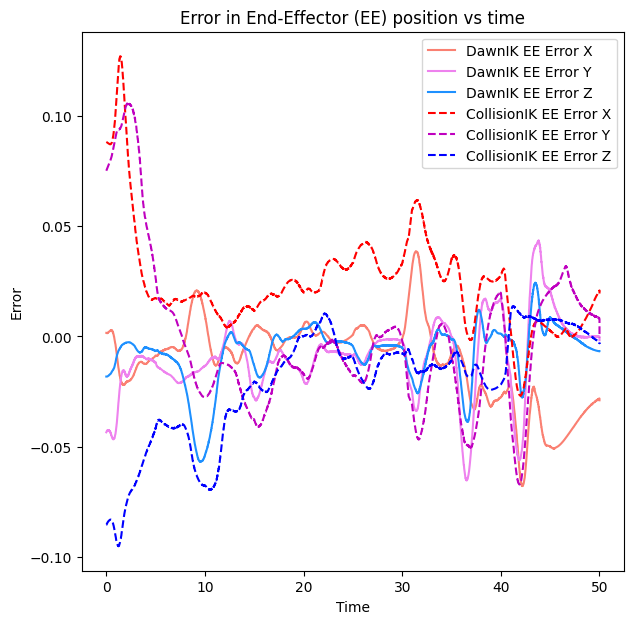} 	
		\captionsetup{width=0.97\columnwidth, justification=justified}
		\caption{Mean error in end-effector positions for DawnIK (dashed lines) and CollisionIK (solid lines) for all the trials in Scenario 2.} 
		\label{fig:s2_ee_error}
	\end{subfigure}
	\begin{subfigure}[t]{0.33\linewidth} 		\centering
		\includegraphics[width=\columnwidth]{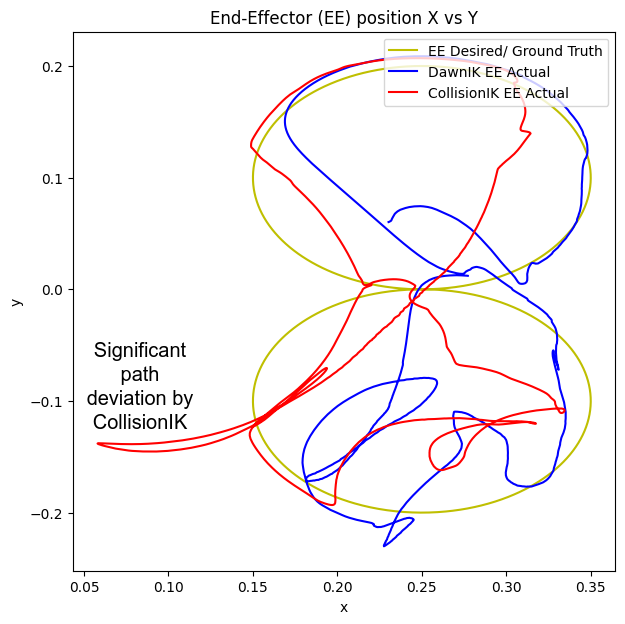} 	
		\captionsetup{width=0.97\columnwidth, justification=justified}
		\caption{Mean path traced by the end-effector using DawnIK (blue) and CollisionIK (red) while drawing an eight in the X-Y plane for Scenario 2.} 
		\label{fig:s2_eight_xy}
	\end{subfigure}
	\begin{subfigure}[t]{0.33\linewidth} 		\centering
		\includegraphics[width=\columnwidth]{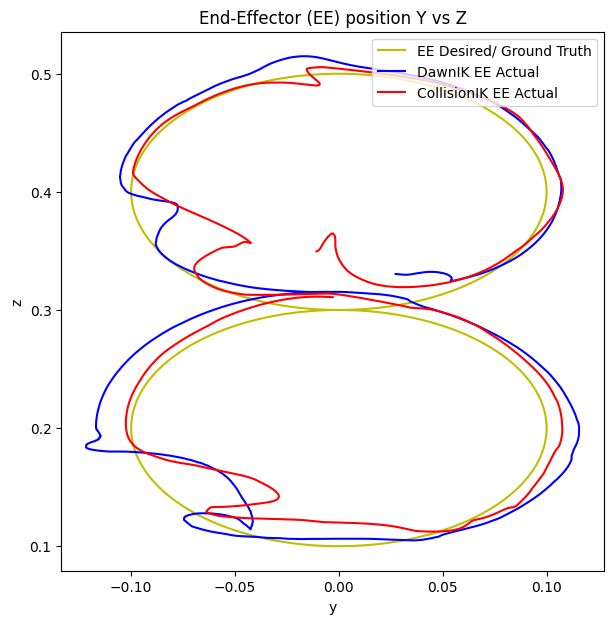} 	
		\captionsetup{width=0.97\columnwidth, justification=justified}
		\caption{Mean path traced by the end-effector using DawnIK (blue) and CollisionIK (red) while drawing an eight in the Y-Z plane for Scenario 2.} 
		\label{fig:s2_eight_yz}
	\end{subfigure}
	\caption{Comparison of performance of DawnIK and CollisionIK in Scenario 2. \figref{fig:s2_ee_error} show that DawnIK leads to lower end-effector position errors over time. \figref{fig:s2_eight_yz} plots provide a visual demonstration of DawnIK's superior trajectory tracking.}
	\captionsetup{justification=justified} 
	\label{fig:scenario_2_plots}
\end{figure*}
In the second scenario, the external Xarm7 arm is repeatedly performing a motion along the X axis, whereas the controlled arm traces a path.
We reuse the paths from Scenario 1. 
In Scenario 2, the external arm repeatedly interferes with the path of the controlled arm, thus causing both the IK solvers to deviate from the desired path.

Both the solvers are successful in avoiding collisions with themselves and with the external arm. 
However, as can been seen from \tabref{tab:results}, the error in end-effector position tracking is higher for CollisionIK as compared to that for DawnIK. 
\figref{fig:s2_eight_xy} also demonstrates that CollisionIK has the tendency to wander further away from the desired trajectory than to avoid the external collisions by moving within a local region. 
\subsubsection{Scenario 3}
\begin{figure*}[t]
	\begin{subfigure}[t]{0.33\linewidth} 		\centering
		\includegraphics[width=\columnwidth]{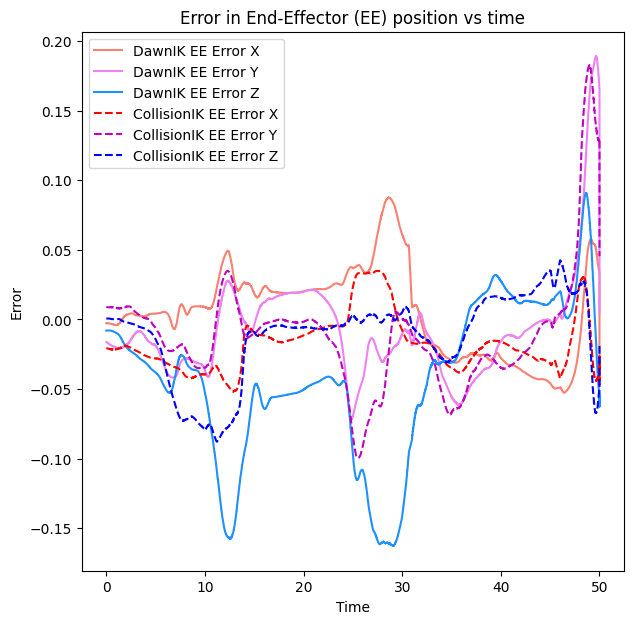} 	
		\captionsetup{width=0.97\columnwidth, justification=justified}
		\caption{Mean error in end-effector positions for DawnIK (dashed lines) and CollisionIK (solid lines) for all the trials in Scenario 3.} 
		\label{fig:s3_ee_error}
	\end{subfigure}
	\begin{subfigure}[t]{0.33\linewidth} 		\centering
		\includegraphics[width=\columnwidth]{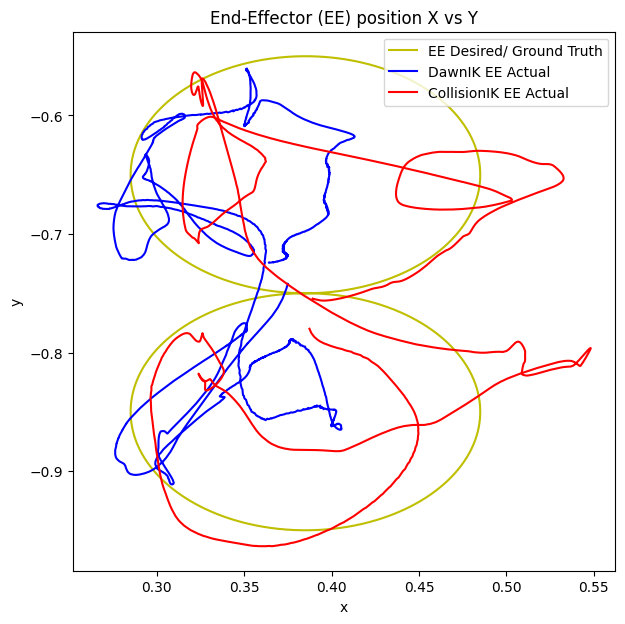} 	
		\captionsetup{width=0.97\columnwidth, justification=justified}
		\caption{Mean path traced by the end-effector using DawnIK (blue) and CollisionIK (red) while drawing an eight in the X-Y plane for Scenario 3.} 
		\label{fig:s3_eight_xy}
	\end{subfigure}
	\begin{subfigure}[t]{0.33\linewidth} 		\centering
		\includegraphics[width=\columnwidth]{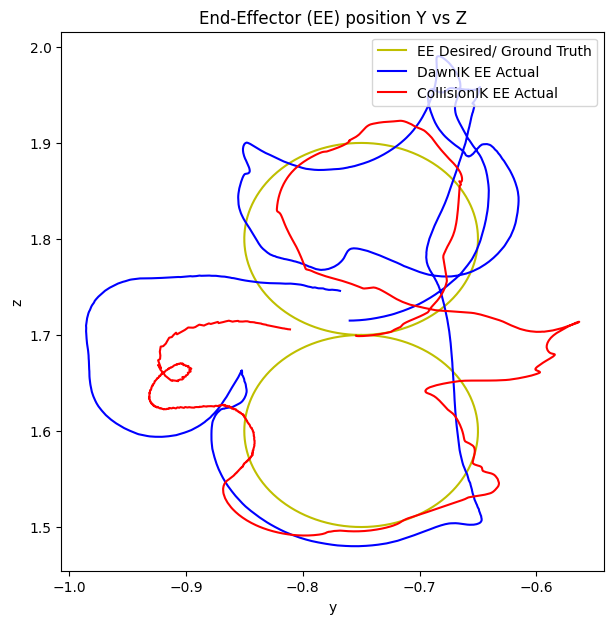} 	
		\captionsetup{width=0.97\columnwidth, justification=justified}
		\caption{Mean path traced by the end-effector using DawnIK (blue) and CollisionIK (red) while drawing an eight in the Y-Z plane for Scenario 3.} 
		\label{fig:s3_eight_yz}
	\end{subfigure}
	\caption{Comparison of performance of DawnIK and CollisionIK in Scenario 3}
	\captionsetup{justification=justified} 
	\label{fig:scenario_3_plots}
\end{figure*}
In the third scenario with the HortiBot system, we omit the square paths as a major portion of the square paths intersect with the stationary arms. 
Instead, we consider a stationary pose for the controlled arm, and also trace an object following path. 
For the external dual arms, we simulate a trajectory where the left arm moves to grasp a fruit while the right arm moves to cut the fruit. 
The grasping arm then moves to drop the object in the front and then the two arms move to their initial poses. 

The configuration of the three arm system and the trajectory of the external dual arms make it extremely challenging for the controlled arm to follow its path. 
With the conflicting constraints of trajectory following and collision avoidance, DawnIK successfully avoids collisions in all the trials. However, this comes at the cost of increased trajectory tracking error. CollisionIK, on the other hand, shows trajectory tracking error similar to that in Scenario 2. 
This leads to multiple states where collisions occur between the controlled arm and the dual arms. 
The offline collision checking with actual mesh models detected a large number of such collisions. 
This shows, that DawnIK is better suited to avoiding collisions with multiple arms. We also validated our method for scenario 3 in the real world using the HortiBot system as can be seen in the accompanying video\footnote{https://youtu.be/-k7XJkbAB6A}.  
\subsubsection{Scenario 4}
\begin{table*}[t] \centering
	\renewcommand{\arraystretch}{1.4}
	\begin{tabular}{|c|c|c|c|c|c|c|c|c|}\hline
		& x (mm)          & y (mm) & z (mm) & roll ($10^{-3}$rad) & pitch ($10^{-3}$rad) & yaw ($10^{-3}$rad)  & collisions\\ \hline
		S4-Xarm7      &17.15 $\pm$ 18.99               &21.21 $\pm$ 27.34  &24.67 $\pm$ 38.34  &6.98 $\pm$ 53.41  &0.99 $\pm$ 1.78  &6.64 $\pm$ 52.91  &0\\ \hline
		S4-Lite6 &23.08 $\pm$ 29.23   &29.30 $\pm$ 26.71  &28.56 $\pm$ 39.57  &1.27 $\pm$ 2.13  &0.44 $\pm$ 2.61  & 1.59 $\pm$ 1.65 &0\\ \hline 
	\end{tabular}
	\captionsetup{width=0.97\linewidth, justification=justified}
	\caption{Mean end-effector position and orientation errors, and mean number of collisions for Lite6 and Xarm7 for Scenario 4. As can been seen, even with the paths intersecting each other, no collisions occur for both Lite6 and Xarm7 arms. While in Scenario 2, the external Xarm7 was performing a trajectory beside the Lite6 arm, in this scenario, they are both occupying the same workspace. Inspite of that, the trajectory tracking errors are comparable showing the benefit of multi-arm collision-aware approach.}
	\label{tab:s4_results}
\end{table*}
\begin{figure*}[t]
	\begin{subfigure}[t]{0.33\linewidth} 		\centering
		\includegraphics[width=\columnwidth]{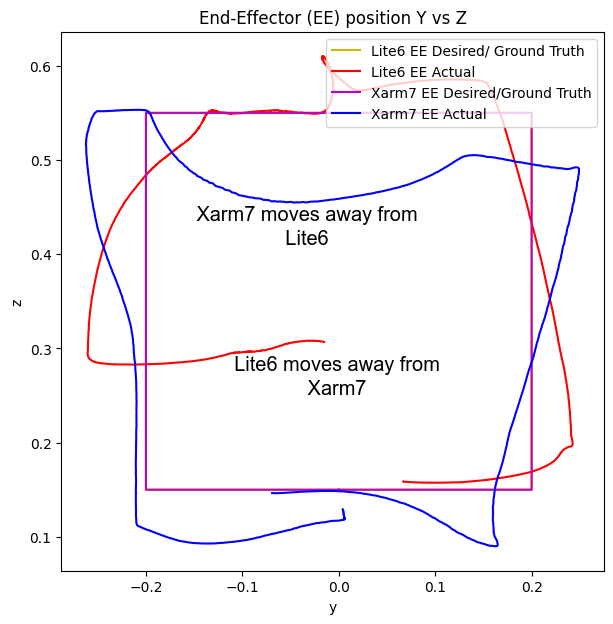} 	
		\captionsetup{width=0.97\columnwidth, justification=justified}
		\caption{Paths traced by Lite6 and Xarm7 while drawing the same square in Y-Z plane for Scenario 4.} 
		\label{fig:s4_square_yz}
	\end{subfigure}
	\begin{subfigure}[t]{0.33\linewidth} 		\centering
		\includegraphics[width=\columnwidth]{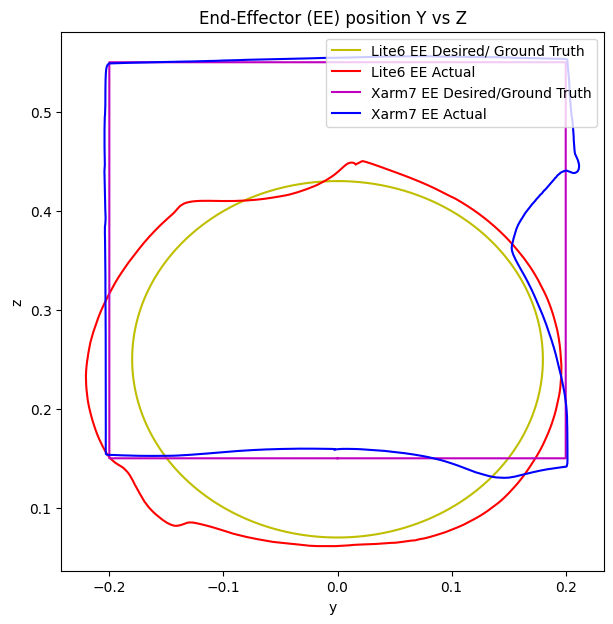} 	
		\captionsetup{width=0.97\columnwidth, justification=justified}
		\caption{Paths traced by Lite6 which draws a circle in Y-Z plane and  Xarm7 draws square in Y-Z plane for Scenario 4.} 
		\label{fig:s4_circle_yz}
	\end{subfigure}
	\begin{subfigure}[t]{0.33\linewidth} 		\centering
		\includegraphics[width=\columnwidth]{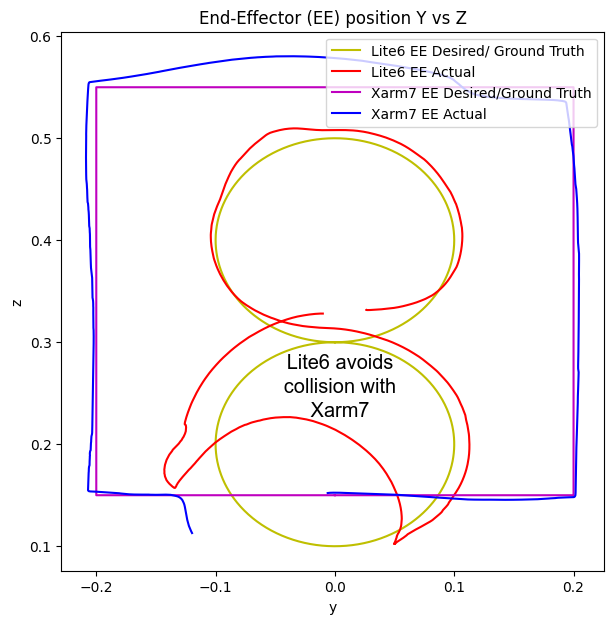} 	
		\captionsetup{width=0.97\columnwidth, justification=justified}
		\caption{Paths traced by Lite6 which draws figure eight in Y-Z plane and  Xarm7 draws square in Y-Z plane for Scenario 4.} 
		\label{fig:s4_eight_yz}
	\end{subfigure}
	\caption{Comparison of performance of DawnIK and CollisionIK in Scenario 4. The Xarm7 performs tracing of the square in the Y-Z plane in all three figures. In \figref{fig:s4_square_yz}, the Lite6 arm traces the same square. The two arms not only do not collide with each other, but they also try to follow the square path to the maximum extent possible. In \figref{fig:s4_circle_yz} and \figref{fig:s4_eight_yz}, Xarm7 demonstrates better tracking as DawnIK is implicitly able to use the redundancy for collision avoidance.}
	\captionsetup{justification=justified} 
	\label{fig:scenario_4_plots}
\end{figure*}

In Scenario 4, both the Lite 6 and the Xarm7 arms are independently controlled by DawnIK. Unlike Scenario 2, where the Xarm7 was performing a point-to-point motion beside the Lite 6 arm, in Scenario 4, it is always tracing the square path in the Y-Z plane. 
Additionally, the start positions are also quite close to each other.
The Lite 6 arm traces the same paths as in Scenario 2, except for the circle in the X-Y plane as it is infeasible due to the Xarm7's path. 

As can been seen in \tabref{tab:s4_results}, no collisions are reported even though the trajectories of both arms intersect multiple times. Both the arms are also able to track the end-effector positions with errors similar to Scenario 2. The 7 DoF Xarm7 has a better trajectory tracking performance as DawnIK implicitly makes use of the redundancy to find feasible solutions. 
\textcolor{black}{The trajectory tracking performance of xArm7 using DawnIK shows that while we do not explicitly exploit redundancy using recursive Jacobian implementations \cite{slotine1991general,zhao2021solving}, our method is still able to use the redundancy to avoid collisions while tracking the trajectory as close as possible.}
\figref{fig:s4_circle_yz} and \figref{fig:s4_eight_yz} show that with the decentralized approach, the two arms are able to track trajectories even in the same plane without any collisions. 
In \figref{fig:s4_square_yz}, the two arms have an extremely challenging task of following the same path while avoiding collisions with each other. However, no collision was detected and both arms are able to follow the square path where feasible. In all the four scenarios, the controlled arm does not get stuck in any singularity for both the solvers. 

\section{Summary}
\label{sec:concl}
With multiple robotic systems working in close proximity, decentralized collision avoidance is a major challenge in manipulator motion planning.
In this work, we developed a decentralized collision-aware inverse kinematics solver that uses trust region based non-linear optimization to follow end-effector trajectories while avoiding collisions. 
We developed a fast and computationally efficient collision distance computation method. 
With only the current joint positions of the external arms and the robot model exposed, DawnIK can successfully avoid collisions in challenging scenarios. 
The results show that DawnIK is better than a state-of-the-art inverse kinematic solver at balancing between trajectory tracking and collision avoidance.
In scenarios where the probability of collision is low, DawnIK shows better performance at trajectory tracking. 
On the other hand, in scenarios where collision is imminent, DawnIK prioritizes collision avoidance. 
Finally, we also demonstrated that we can enable independent yet simultaneous control of multiple arms using DawnIK solver for trajectory tracking while avoiding collisions with each other efficiently.

\bibliographystyle{IEEEtran}

\balance 
\bibliography{refs}
\balance 

\end{document}